\documentclass[conference,a4paper]{APSIPA2025}
\usepackage{amsmath}
\usepackage{graphicx}
\usepackage{amssymb}
\usepackage{multirow}
\usepackage{makecell}
\usepackage{booktabs}
\usepackage{pifont}
\usepackage{threeparttable}
\usepackage[backend=biber,style=ieee,]{biblatex}
\usepackage{geometry}
\usepackage{fancyhdr}

\fancypagestyle{firststyle}{
  \fancyhf{}
  \fancyhead[C]{2025 Asia Pacific Signal and Information Processing Association Annual Summit and Conference (APSIPA ASC)}
}

\begin{document}

% \title{Contextual ASR Enhancement via Synthesis-Driven Multi-Pronunciation Hotword Prefix-Trie}
% \title{Synthesis-Driven Multi-Pronunciation Prefix-Trie for Zero-Shot Contextual ASR}
\title{Zero-shot Context Biasing with Trie-based Decoding using Synthetic Multi-Pronunciation}
% \title{Zero-Shot Contextual ASR Using Multi-Pronunciation Hotword for Whisper}
% \title{Multi-Pronunciation Hotword for Zero-Shot Whisper Contextual ASR}
% \title{Contextual ASR Enhancement via TTS-Generated Multipronunciation Hotwords
% }

\author{
\authorblockN{
Changsong Liu\authorrefmark{1}\authorrefmark{2},
Yizhou Peng\authorrefmark{1}\authorrefmark{2}, and Eng Siong Chng\authorrefmark{2}
}

% \authorblockA{
% Nanyang Technological University, Singapore \\
% E-mail: changsong.liu@ntu.edu.sg}
% }
%%%
\authorblockA{
\authorrefmark{1}
Alibaba-NTU Global e-Sustainability CorpLab, Nanyang Technological University, Singapore \\
}

\authorblockA{
\authorrefmark{2}
College of Computing and Data Science, Nanyang Technological University, Singapore \\
E-mail: changsong.liu@ntu.edu.sg}
}

\maketitle
\pagestyle{empty}
\thispagestyle{firststyle}

\begin{abstract}
Contextual automatic speech recognition (ASR) systems allow for recognizing out-of-vocabulary (OOV) words, such as named entities or rare words. However, it remains challenging due to limited training data and ambiguous or inconsistent pronunciations. In this paper, we propose a synthesis-driven multi-pronunciation contextual biasing method that performs zero-shot contextual ASR on a pretrained Whisper model.
Specifically, we leverage text-to-speech (TTS) systems to synthesize diverse speech samples containing each target rare word, and then use the pretrained Whisper model to extract multiple predicted pronunciation variants. These variant token sequences are compiled into a prefix-trie, which assigns rewards to beam hypotheses in a shallow-fusion manner during beam-search decoding. Subsequently, any recognized variant is mapped back to the original rare word in the final transcription.
% improves recognition of rare hotwords without requiring any adaptation to the underlying ASR model. 
% Specifically, we synthesize speech samples for each hotword using diverse TTS engines and speaker styles, and transcribe them with Whisper to extract multiple pronunciation variants. These variants are used to construct a multi-pronunciation prefix-trie. At inference time, we incorporate this prefix-trie-based module into Whisper in a shallow fusion manner, assigning a uniform per-token reward to guide recognition of multi-pronunciation hotword vatiants, which are then mapped back to the correct spelling. 
% Evaluated on the LibriSpeech test sets benchmark, 
The evaluation results on the LibriSpeech dataset show that our method reduces biased-word error rate (B-WER) by 43\% on test-clean and 44\% on test-other while maintaining unbiased-WER (U-WER) essentially unchanged.

\end{abstract}


\section{Introduction}
Contextual automatic speech recognition (ASR), also known as “hotword” or "named-entity" customization, enables ASR systems to accurately recognize user-specified terms, such as names, places, or technical jargon, that are rare or even out-of-vocabulary (OOV) in training data~\cite{advancesasr}. Modern end-to-end (E2E) ASR models achieve an impressive overall word error rate (WER) compared to traditional hybrid systems, thanks to advanced architectures such as RNNs~\cite{rnn}, Transformers~\cite{attention}, and Conformers~\cite{conformer}. However, due to data scarcity and confusion in pronunciation for rare words, contextual ASR remains challenging~\cite{seacoparaformer}. 
% but still suffer high error rates on these rare hotwords due to pronunciation mismatch and limited training examples \cite{seacoparaformer}.

Traditional contextual biasing techniques augment decoding via shallow‐fusion with an external language model \cite{shallowfusion} or on-the-fly WFST rescoring \cite{wfstrescoring}. While effective, these methods require careful tuning of fusion weights and can degrade WER/U-WER performance when the bias list becomes large, due to an increase in false positive distractor terms with similar spellings or pronunciations.
To be more straightforward, the E2E approaches embed context phrases directly into the ASR model, for example,  Contextual Listen, Attend and Spell (CLAS) uses an attention‐based contextual module to bias decoding \cite{attentioncontextual}, and trie-based deep biasing integrates dynamic n-gram constraints into transducer decoding \cite{triebiasing}. 
Recently, large language model (LLM)-based ASR has emerged, leveraging prompt engineering to incorporate hotwords into decoding. For example, a CTC-assisted LLM-based contextual ASR model uses coarse CTC hypotheses to select effective hotwords from a large hotword list, which are then fed into an LLM as a prompt~\cite{ctcllm}.
These yield substantial improvements in biased-WER (B-WER) by enhancing the model's capability for contextual understanding during the inference phase.

Despite these advances, one critical factor has often been overlooked: the variability in pronunciation of hotwords. In practice, many hotwords, especially specialized terms such as technical jargon, foreign names, or branded entities, are unfamiliar to users, leading to inconsistent or incorrect pronunciations. As a result, the same hotword may be spoken in multiple ways across different speakers or even by the same speaker over time. This mismatch between the intended hotword and its spoken realization can significantly degrade the effectiveness of contextual biasing, as models may fail to associate these variant forms with the intended target.
% but still struggle with pronunciation variability between hotwords and their spoken forms.
% Despite these advances, no prior work explicitly models multiple pronunciation variants of each hotword, leaving residual errors when ASR tokenization does not match the original form.

% In this paper, we introduce a generalizable TTS-augmented multi-pronunciation pipeline for contextual ASR with a modified Whisper decoder:

% In this paper, we introduce a novel approach to enhancing the recognition accuracy of various pronunciations for hotwords in contextual ASR systems.
% In this paper, we introduce a novel approach to enhancing the recognition accuracy of rare words in contextual ASR systems by leveraging synthetic speech to generate diverse pronunciation variants, without requiring adaptation of the ASR model. 
% In this paper, we introduce a novel approach to enhancing the recognition accuracy of rare words in contextual ASR systems by leveraging synthetic speech generated from multiple TTS engines and diverse speaker styles. A pretrained ASR model transcribes these synthetic utterances to obtain diverse pronunciation variants, all without requiring adaptation of the ASR model.
In this paper, we introduce a novel approach to enhancing the recognition accuracy of rare words in a pretrained ASR system without requiring any adaptation. Our method leverages synthetic speech to generate diverse pronunciation variants for each hotword, which are then transcribed by the same ASR model. 
Technically, we synthesize speech using a variety of TTS systems and voices for each templated hotword sample and apply Whisper Large-v3 model to transcribe them. We integrate a multi-pronunciation prefix-trie-based hotword module into the same Whisper Large-v3 model in a shallow-fusion manner, where the prefix-trie is constructed using extracted multiple pronunciation variants of each hotword. 

Our experiments on the LibriSpeech test sets demonstrate that the proposed method reduces WER/B-WER of the baseline Whisper Large-v3 model without affecting the recognition performance of common words. To the best of our knowledge, this is the first work to leverage synthesized multi-pronunciation variants for contextual ASR without requiring model adaptation.

\section{Methodology}
\label{sec:med}
In this section, we describe our end‐to‐end pipeline for augmenting Whisper’s contextual biasing with TTS‐generated multi-pronunciation variants. The pipeline comprises three sequential stages: Hotword Speech Synthesis, Hotword Token Extraction, and Prefix-Trie Implementation in Whisper.
% hotword variation synthesis, variant token extraction, and trie‐biased decoding.

\begin{figure*}[ht]
  \centering
  \includegraphics[width=1.0 \linewidth]{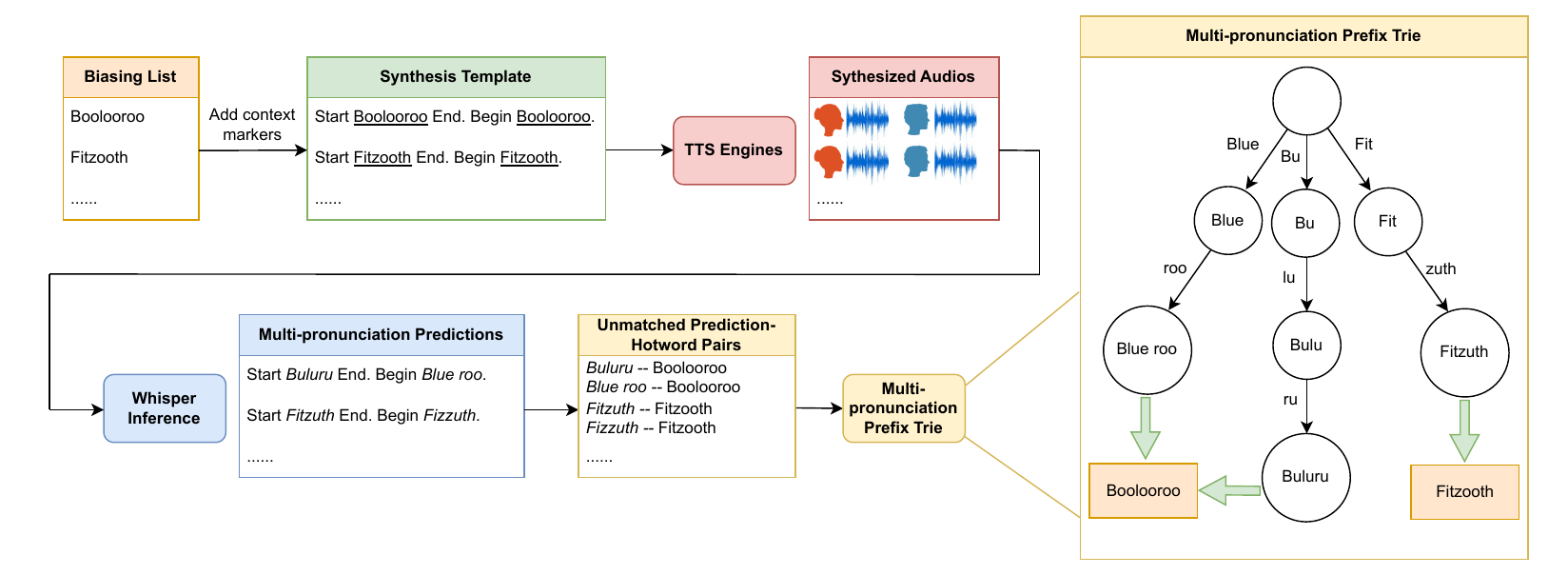}
  % \caption{TTS Synthesis powered Multi-Pronunciation Prefix-Trie for the Whisper model. With a given hotword list, we fill each hotword into the designated synthesis template, which is then fed into several TTS engines to mimic three different speakers' voices to synthesize speech. Then, the speech is recognized using the vanilla Whisper model, and the predictions are used to extract multiple pronunciation variations. Each written form of pronunciation is treated as an effective hotword when building the Prefix-Trie, and it will be replaced with the original hotword when decoding finishes.} 
  \caption{Workflow of TTS-powered multi-pronunciation prefix-trie construction for Whisper model. Given a list of hotwords, each is inserted into a designated template and synthesized using multiple TTS engines to mimic three distinct speaker voices. The synthetic speech is then transcribed using the vanilla Whisper model, from which diverse pronunciation variants are extracted. Each unique transcription is included as an effective hotword in the prefix-trie. During decoding, any recognized variant shown in the final results is mapped back to the original hotword.}
  % \caption{Diagram for TTS-augmented multipronunciation pipeline.}
  \label{fig:diagram}
\end{figure*}

\subsection{Multi-Pronunciation Hotword Speech Synthesis via TTS}

We begin by synthesizing multiple pronunciation variants for each target hotword using three complementary TTS engines, including CosyVoice~\cite{cosyvoice}, F5-TTS~\cite{f5tts}, and GPT-SoVITS~\cite{gpt-sovits}.

\begin{itemize}
    \item \textbf{F5-TTS} is a non-autoregressive (NAR) TTS model that builds on a Diffusion Transformer (DiT) architecture and incorporates ConvNeXt V2 blocks to effectively address text-speech alignment in an in-context learning setting. It is trained on 95K hours of \textbf{English-Chinese} bilingual dataset and supports good-quality synthesis in both languages.
    \item \textbf{CosyVoice} is a scalable zero-shot multilingual TTS system that employs a text-to-token-to-speech architecture. It leverages supervised semantic tokens extracted from a quantized multilingual ASR encoder, uses a large language model (LLM) to predict token sequences from text and speaker embeddings, and reconstructs speech via a conditional flow matching model followed by a HiFi-GAN vocoder. It is trained on around 170K hours of multilingual dataset and supports five languages and dialects, including \textbf{Chinese}, \textbf{English}, \textbf{Cantonese}, \textbf{Japanese}, and \textbf{Korean}.
    \item \textbf{GPT-SoVITS} integrates a speech-to-unit encoder, a speaker encoder, and a GPT-style text-to-unit decoder within the So-VITS framework. It enables prompt-based zero-shot voice cloning by generating discrete unit sequences from text and synthesizing them into a waveform via a neural vocoder. It supports the same five languages as CosyVoice.
\end{itemize}
% F5-TTS is a non-autoregressive (NAR) TTS model  that builds on a Diffusion Transformer (DiT) architecture and incorporates ConvNeXt V2 blocks to effectively address text-speech alignment in an in-context learning setting. It is trained on 95K hours of English-Chinese bilingual dataset and supports good-quality synthesis in both languages.

% CosyVoice is a scalable zero-shot multilingual TTS system that employs a text-to-token-to-speech architecture. It leverages supervised semantic tokens extracted from a quantized multilingual ASR encoder, uses a large language model (LLM) to predict token sequences from text and speaker embeddings, and reconstructs speech via a conditional flow matching model followed by a HiFi-GAN vocoder. It is trained on around 170K hours of multilingual dataset and supports 5 languages and dialects including Chinese, English, Cantonese, Japanese, and Korean.

% GPT-SoVITS integrates a speech-to-unit encoder, a speaker encoder, and a GPT-style text-to-unit decoder within the So-VITS framework. It enables prompt-based zero-shot voice cloning by generating discrete unit sequences from text and synthesizing them into waveform via a neural vocoder. It supports the same 5 languages as Cosyvoice.

All three systems support prompt-based speech synthesis by conditioning on a reference audio, its corresponding text, and a new target text, allowing for zero-shot speech generation in the reference speaker’s voice and style.

To ensure that the hotword appears in a fixed, easily locatable context, we embed it in two minimal utterances: “Start \textless{}Hotword\textgreater{} End” and “Begin \textless{}Hotword\textgreater{}”. The context can be customized, but we adopt such simple and rigid templates to facilitate reliable extraction from Whisper decoding results and to ensure stable and consistent synthesis from TTS models. Longer or more complex contexts tend to introduce unwanted variability in TTS outputs, which can compromise the reliability of anchor-based hotword localization, as the surrounding context serves as a temporal marker for extracting hotword speech variations. Each engine renders both utterances with three distinct voices and accents from~\cite{voiceover_samples}, including British female, British male, and American male. This process yields a total of 3 engines × 2 utterances × 3 voices = 18 synthesized utterances per hotword, capturing a rich and comprehensive variety of pronunciation patterns.

\subsection{Multi-Pronunciation Hotword Token Extraction}

%%%%%%%%%%%%%%%%%%%%%%%%%%%%
\begin{figure*}[ht]
  \centering
  \includegraphics[width=0.9 \linewidth]{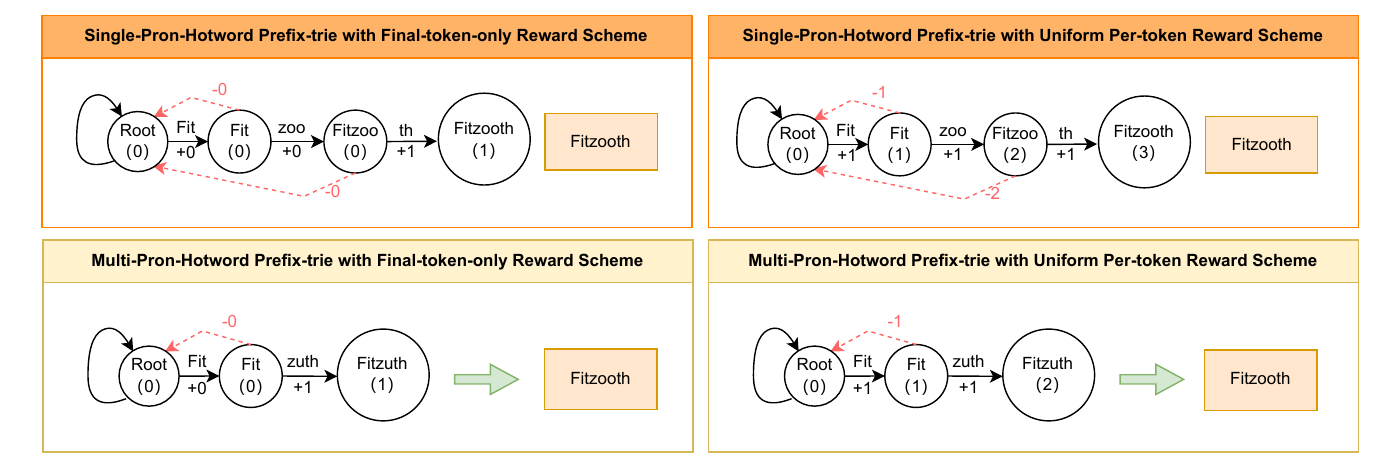}
  \caption{An illustration of Single-Pron-Hotword and Multi-Pron-Hotword prefix-trie using \textbf{Final}-token-only and \textbf{Uniform} per-token reward schemes. In Single-Pron-Hotword prefix-trie, each hotword corresponds to only one path, representing the correct spelling. In contrast, in a Multi-Pron-Hotword prefix-trie, each hotword may correspond to multiple paths with different spellings. When using the \textbf{Final} reward, only the last state of a valid path will give a reward of 1; whereas when using \textbf{Uniform} reward, each active state will credit a reward of 1 when matched. Numbers in the parentheses stand for the rewards gained up to the current state. Red dashed lines are examples of failing to match in the middle state of a prefix-trie path.}
  % \caption{Diagram for TTS-augmented multipronunciation pipeline.}
  \label{fig:diagram2}
\end{figure*}
%%%%%%%%%%%%%%%%%%%%%%%%%%%%
Next, each synthesized waveform is fed into the vanilla Whisper Large-v3 model to obtain the recognition results, where we require not only the final sentences but also the sequences of byte-pair encoding (BPE) tokens. This is to align with the requirement of prefix-trie as mentioned in Section~\ref{ssec: trie-biasing}, which performs hotword biasing through BPE-level rewarding scores in each auto-regressive decoding step of the Whisper decoder that generates the probability distribution of the entire BPE dictionary.

From the token sequences, we locate the tokens corresponding to the known context markers “Start,” “End,” and “Begin,” and we isolate the subsequences that occur between “Start” and “End” as well as those immediately following “Begin”. Any extracted sequence whose token IDs exactly match the original hotword is discarded, so that only unmatched pronunciation-prediction variant pairs remain. This extends the search space to include more predictions that the Whisper model can make for each hotword, allowing us to improve hit accuracy.
% Additionally, we refine the variant set to eliminate faulty options by counting the number of phonemes (syllables).
Additionally, we refine the variant set to eliminate faulty options by Syllable Filtering (SF). 
Specifically, we use the syllapy~\cite{syllapy} package to estimate the number of phonemes of the predicted variants. Only those with syllable count matches that of the original hotword and at least three are retained. This strategy significantly improves performance by removing inappropriate pronunciation variants introduced by TTS generation instability or the Whisper model's hallucinations through the automated pipeline.
% each token sequence is evaluated for syllable count using a phoneme‐based heuristic; only variants whose syllable count matches that of the original hotword and is at least three are retained.
% The necessity of syllable filtering is demonstrated in an ablation study.

\subsection{Prefix-Trie Implementation in Whisper Decoder}
\label{ssec: trie-biasing}

% We begin by unifying the original hotword token sequence and all TTS‐generated variants into a single finite‐state trie.
We implement a simple prefix-trie module into the Whisper beam-search decoding to enable contextual biasing in Whisper without requiring model adaptation or other learnable modules, such as Tree-constrained Pointer Generator (TCPGen)~\cite{TCPGen}. 
In our prefix-trie architecture, each arc, corresponding to a token, is annotated with a Uniform per‐token reward of \(+1\), so that every token in every pronunciation variant contributes equally to decoding. Since each token in a hotword contributes a fixed reward during decoding, longer hotwords (i.e., those with more tokens) inherently receive more cumulative bias, giving them a greater advantage when matched correctly. During beam‐search decoding, we maintain, for each beam hypothesis, a corresponding “active” trie state that records how far along any hotword prefix that hypothesis has traversed. 
Given the audio $X$, at each time step \(t\), when considering an output token \(y_{t}\), we first compute the \textbf{W}hisper score $S_{\textbf{W}}$ as in Eq.~\ref{eq:model_score}.  
\begin{equation}
    \label{eq:model_score}
    {S_{\textbf{W}}}(y_{t}) = - \log P_{\textbf{W}}(y_{t}\mid y_{t-1},..., y_{0}, X)
\end{equation}
Here we define the cumulative score of each beam \textbf{H}ypothesis as in Eq.~\ref{eq:hyp_cumulative_score}.
\begin{equation}
    \label{eq:hyp_cumulative_score}
    {S_{\textbf{H}}}(y_{t}) = {S_{\textbf{H}}}(y_{t-1}) + {S_{\textbf{W}}}(y_{t}) 
\end{equation}
We then attempt to advance the hypothesis’s current trie state with \(y_{t}\): if \(y_{t}\) continues a valid hotword prefix conditioned on $y_{t-1}$, the trie returns a reward \(\rho_{y_{t-1} \rightarrow y_{t}}=1\); otherwise we remove the accumulated reward of partial unsuccessful match from the hypothesis's cumulative score. We define the reward as \(- \sum_{i=1}^{n} \rho_{y_{t-i} \rightarrow y_{t-i+1}}\) where \(n\) is the depth from root to current state of prefix-trie as shown in Figure~\ref{fig:diagram2}. We form the final score $S$ augmented with contextual rewards, as in Eq.~\ref{eq:final_score}, which is then used in place of ${S_{\textbf{W}}}$ when ranking and pruning beam candidates.
\begin{equation}
    \label{eq:final_score}
    S(y_{t}) = {S_{\textbf{W}}}(y_{t}) + \rho_{y_{t-1} \rightarrow y_{t}} 
\end{equation}
% \[
%    S_i(t)\;=\;\log P_{\mathrm{model}}(t\mid \text{prefix}_{i-1})\;+\;\rho_i
% \]
% \(S\) is then used instead of the raw model score when ranking and pruning beam candidates.
In this way, any hypothesis that follows one of our multi‐pronunciation hotword paths receives a consistent positive boost, biasing the search toward recognizing those variants. After the beam-search decoding, % end‐of‐text token is generated, 
we select the highest‐scoring beam, and then traverse its recorded trie states to map to the correct hotword spelling. Since we only add these Uniform per-token rewards to the beam scores without altering the underlying acoustic or language model logits, our trie‐based biasing improves recognition of the target hotwords while preserving Whisper’s overall decoding behavior and general ASR accuracy.  
Through this three-stage methodology, as shown in Figure~\ref{fig:diagram}, we incorporate a prefix-trie-based multiple pronunciation hotword mechanism with Whisper for contextual biasing, leading to significant reductions in B-WER without degrading overall ASR accuracy.

\section{Experiments}

% In this section, we evaluate our TTS-augmented multipronunciation biasing pipeline on the LibriSpeech benchmark. We first describe the dataset, evaluation metrics, and implementation details. We then compare our proposed method with different configurations. Next, we isolate the contributions of syllable filtering and trie reward assignment in ablation studies. Finally, we compare our best system against leading contextual ASR methods, demonstrating that with no fine-tuning or external LLM, we outperform many finetuned models and approach the state of the art. 
In this section, we introduce the dataset and evaluation metrics used, as well as the experimental setup.

\subsection{Dataset and Evaluation Metrics}
\subsubsection{Data and Hotword Selection}
We follow previous research~\cite{triebiasing} on their settings regarding contextual biasing and benchmark our method on the widely used LibriSpeech dataset~\cite {librispeech}, including the \textit{test-clean} and \textit{test-other} evaluation sets. 
% Specifically, we adopt the artificial biasing list setup from \cite{triebiasing}. In this approach, the 5,000 most frequent words from the LibriSpeech training data are labeled as \textit{common}, while all remaining words are treated as \textit{rare}. For each test utterance, the biasing list is composed of two parts: rare words that appear in the reference transcription, and distractor words randomly selected from the rare word vocabulary of 209.2K entries. To evaluate robustness under different levels of distraction, lists of varying sizes are constructed by having N = {100, 500, 1000, 2000} distractors.
Specifically, we adopt the artificial biasing word list setup\footnote{%
  \raggedright
  The rare and common word lists are from \url{https://github.com/facebookresearch/fbai-speech/tree/main/is21_deep_bias}.
} where the 5,000 most frequent words from the LibriSpeech training data are labeled as \textit{common}, while all remaining words are treated as \textit{rare}, including over 209.2K words. For each test utterance, the biasing list is composed of two components: rare words that appear in the reference transcription, and distractor words randomly selected from the \textit{rare} words list. We use a fixed $N = 1000$ distractors to align with prior works for a fair comparison.
The statistics of both test sets are as demonstrated in Table \ref{tab:bias_stats}. Rare words take up 10\% of the word count in both test sets.
% To evaluate robustness under different levels of distraction, lists of varying sizes are constructed by having N = {100, 500, 1000, 2000} distractors.

From the rare word vocabulary, we select a subset of rare words that exhibit a non-zero initial WER when decoded with the official Whisper Large-v3 model without contextual biasing. We then perform speech synthesis on this subset. In contrast, rare words that are correctly transcribed are explicitly excluded from further processing and do not go through the synthesis pipeline. Both sections of rare word vocabulary above are taken into account when calculating B-WER.

%%%%%%%%%%%%%%%%%%%%%%%%%%%%%%%
\begin{table}[ht]
\centering
\caption{Statistics of the LibriSpeech dataset. Appearing rare words refers to the rare word counts that are present in the reference transcripts.}
\label{tab:bias_stats}
\begin{tabular}{lcc}
\toprule
\textbf{Metric} & \textbf{test-clean} & \textbf{test-other} \\
\midrule
Total words & 52,576 & 52,343 \\
Common words & 46,815 & 46,993 \\
Appearing rare words & 5,761 & 5,350 \\
Utterances & 2,620 & 2,930 \\
\midrule
Avg. rare words per utterance & 2.20 & 1.83 \\
Rare word rate & 10.96\% & 10.22\% \\
\bottomrule
\end{tabular}
\end{table}

%%%%%%%%%%%%%%%%%%%%%%%%%%%%%%%%%%%%%

\subsubsection{Evaluation Metrics}
We utilize commonly used evaluation metrics for contextual ASR tasks: overall WER to measure general recognition quality, biased-WER (B-WER) to specifically evaluate the accuracy on targeted hotwords, and unbiased-WER (U-WER) to measure the accuracy on common words.
Comparing contextual ASR with standard ASR systems, the combination of these three WERs shows the effectiveness (true positives) and distractions (false positives) of contextual biasing strategies embedded in the ASR systems. Thus, an optimal contextual biasing ASR should achieve improved B-WER while maintaining U-WER unchanged.
For transcription normalization and metrics calculation, we utilize the scripts provided by MetaAI~\cite{triebiasing}.
% to ensure that hotword biasing does not negatively impact the recognition of common vocabulary. 
% The transcripts are normalized, and the evaluation metrics are computed with scoring scripts from \cite{triebiasing}.

\subsection{Experimental Setup}
\label{ssec:experiment_setup}
All of our experiments are based on the Whisper Large-v3 model, and we use beam-search decoding with a beam size of 10. 
We use the vanilla Whisper Large-v3 model as a baseline, comparing against the Whisper Large-v3 with prefix-trie hotword module under various configurations as illustrated in Figure \ref{fig:diagram2}:
\begin{itemize}
    \item \textbf{Single-Pron-Hotword}: Each hotword is bound with only one way of spelling, which is the original correct one, and corresponds to only one path in the prefix-trie.
    \item \textbf{Multi-Pron-Hotword}: As proposed in Section~\ref{sec:med}, each hotword is bound with several selected pronunciations generated through a TTS-Whisper pipeline, and may correspond to several paths in the prefix-trie.
\end{itemize}
% Furthermore, we explore two reward strategies in the Prefix-Trie: a \textbf{Uniform} per-token reward scheme and a \textbf{Final}-token-only reward scheme. If a valid hotword prefix trie path expands from token step \(1\) to \(t\), the uniform per-token reward scheme will return \(\rho_{y_{1} \rightarrow y_{2}}=\rho_{y_{2} \rightarrow y_{3}}=...=\rho_{y_{t-1} \rightarrow y_{t}}=1\). In contrast, the final-token-only reward scheme will return \(\rho_{y_{1} \rightarrow y_{2}}=\rho_{y_{2} \rightarrow y_{3}}=...=\rho_{y_{t-2} \rightarrow y_{t-1}}=0\) and \(\rho_{y_{t-1} \rightarrow y_{t}}=1\).
Furthermore, we explore two reward strategies in the prefix-trie: a \textbf{Final}-token-only reward scheme that gives a reward of $1$ only when the decoding beam reaches the terminal node of the hotword in the trie, and a \textbf{Uniform} per-token reward scheme where each hit token of the hotword returns a reward of $1$. The \textbf{Final} scheme only rewards those beams where Whisper predicts the entire correct sequence of hotwords on its own. In contrast, the \textbf{Uniform} scheme continuously rewards a beam each time a partial token of a hotword is predicted; however, if at any point the next expected token does not match, the awarded reward is immediately removed.
% \begin{itemize}
%     \item \textbf{Uniform per-token reward}: If a valid hotword prefix trie path expands from token step \(1\) to \(t\), the uniform per-token reward scheme will return \(\rho_{y_{1} \rightarrow y_{2}}=\rho_{y_{2} \rightarrow y_{3}}=...=\rho_{y_{t-1} \rightarrow y_{t}}=1\).
%     \item \textbf{Final-token-only reward}: In contrast, the final-token-only reward scheme will return \(\rho_{y_{1} \rightarrow y_{2}}=\rho_{y_{2} \rightarrow y_{3}}=...=\rho_{y_{t-2} \rightarrow y_{t-1}}=0\) and \(\rho_{y_{t-1} \rightarrow y_{t}}=1\).
% \end{itemize}
% We systematically investigate different configurations involving original hotword biasing (self-bias), TTS-generated multipronunciation variants (TTS-bias), and their combination (TTS-hotword-bias). Furthermore, we explore two reward strategies in trie-based decoding: a uniform per-token reward scheme and a final-token-only reward scheme. We also perform ablation studies on the necessity of filtering TTS-generated variants by syllable count to prevent performance degradation due to noisy variants. 

\section{Results}

\begin{table*}[t]
\centering
\caption{Performance of the proposed multi-pronunciation hotword method on LibriSpeech dataset using biasing list with $N=1000$. \textbf{SF} indicates whether we apply syllable filtering to the multi-pronunciation variants. \textbf{Reward} indicates the reward patterns (Final-token-only vs. Uniform per-token) used. Final-token-only indicates only the final prefix-trie state of a valid path has a reward of 1 while the previous states have a reward of 0. Uniform per-token indicates every prefix-trie state of a valid path has a reward of 1. WER refers to the overall word error rate. B-WER/U-WER refers to biased/unbiased-WER, indicating the model's ability to recognize rare/common words. }
\label{tab:proposed_librispeech}
\begin{tabular}{llclcccccc}
\toprule
\multirow{2}{*}{\textbf{System}} &
\multirow{2}{*}{\textbf{Condition}} & \multirow{2}{*}{\textbf{SF}} & \multirow{2}{*}{\textbf{Reward}} & \multicolumn{3}{c}{\textbf{test-clean}} & \multicolumn{3}{c}{\textbf{test-other}} \\
\cmidrule(lr){5-7} \cmidrule(lr){6-10}
 & & & & \textbf{WER} & \textbf{U-WER} & \textbf{B-WER} & \textbf{WER} & \textbf{U-WER} & \textbf{B-WER} \\
\midrule
Baseline & Vanilla Whisper Large-v3 & – & - & 2.81 & 1.94 & 9.86 & 4.84 & 3.39 & 17.55 \\
S1 & Single-Pron-Hotword & – & Final & 2.57 & 1.91 & 7.95 & 4.27 & 3.21 & 13.61 \\
S2 & Multi-Pron-Hotword & \ding{55} & Final & 19.97 & 21.19 & 9.98 & 26.45 & 27.25 & 19.42 \\
S3 & ~~~ + Single-Pron & \ding{55} & Final & 19.79 & 21.17 & 8.54 & 26.15 & 27.21 & 16.84 \\
S4 & Multi-Pron-Hotword & \checkmark & Final & 2.70 & 1.96 & 8.73 & 4.52 & 3.25 & 15.68 \\
S5 & ~~~ \textbf{+ Single-Pron} & \checkmark & Final & \underline{2.50} & \underline{1.92} & \underline{7.19} & \underline{4.18} & \underline{3.22} & \underline{12.60} \\
\midrule
S6 & Single-Pron-Hotword & - & Uniform & 2.33 & \textbf{1.87} & 6.11 & 3.87 & \textbf{3.11} & 10.52 \\
% S4 & Multi-Pron-Hotword & Final & 2.70 & 1.96 & 8.73 & 4.52 & 3.25 & 15.68 \\
% S7 & Multi-Pron-Hotword & Uniform & 2.67 & 1.96 & 8.42 & 4.53 & 3.27 & 15.59 \\
% S5 & ~~~ + Single-Pron & Final & 2.50 & 1.92 & 7.19 & 4.18 & 3.22 & 12.60 \\
S7 & ~~~ \textbf{+ Multi-Pron} & \checkmark & Uniform & \textbf{2.31} & 1.90 & \textbf{5.66} & \textbf{3.83} & 3.14 & \textbf{9.90} \\
\bottomrule
\end{tabular}
\end{table*}

%%%%%%%%%%%%%%%%%%%%%%%%%%%%%%%

\begin{table*}[h]
\centering
\caption{Comparison with prior contextual ASR models on LibriSpeech dataset using biasing list \cite{triebiasing} with list size $N=1000$.}
\label{tab:comparison_librispeech}
\begin{tabular}{lcccccc}
\toprule
\textbf{Model} & \multicolumn{3}{c}{\textbf{test-clean}} & \multicolumn{3}{c}{\textbf{test-other}} \\
\cmidrule(lr){2-4} \cmidrule(lr){5-7}
 & \textbf{WER} & \textbf{U-WER} & \textbf{B-WER} & \textbf{WER} & \textbf{U-WER} & \textbf{B-WER} \\
\midrule
CPPNet~\cite{cppnet} & 3.81 & 2.90 & 11.40 & 8.75 & 6.90 & 25.30 \\
Deep Biasing+BPB~\cite{biasphraseboosted} & 3.47 & 3.00 & 7.70 & 7.34 & 6.40 & 15.80 \\
Whisper+TCPGen+GPT-2~\cite{whisper_gpt2} & 3.40 & -- & 8.20 & 6.30 & -- & 16.30 \\
TCPGen+GNN enc.~\cite{tcpgengnn} & 3.10 & -- & 6.70 & 7.90 & -- & 17.80 \\
GA-CTC~\cite{gactc} & 2.40 & 2.00 & 6.30 & 6.20 & 5.20 & 15.20 \\
TCPGen+p+phn-aware Q~\cite{phoneme} & 2.20 & -- & 4.60 & 6.00 & -- & 12.30 \\
DB-NNLM~\cite{triebiasing} & 2.14 & 1.60 & 6.70 & 6.35 & 5.10 & 17.20 \\
CTC+LLM~\cite{ctcllm} & 1.33 & 1.00 & 4.16 & 2.99 & 2.31 & 9.33 \\
\textbf{S7} & \underline{2.31} & \underline{1.90} & \underline{5.66} & \underline{3.83} & \underline{3.14} & \underline{9.90} \\
\bottomrule
\end{tabular}
\end{table*}

%%%%%%%%%%%%%%%%%%%%%%%%%%%%%%%%%%%%%%%%%%%

% \subsection{Main Results}
Table \ref{tab:proposed_librispeech} summarizes our evaluation results on LibriSpeech test-clean and test-other subsets. 
Firstly, we test the \textbf{Final} reward scheme to evaluate Whisper's ability to recognize rare words, as we only reward hypotheses that contain correctly recognized hotwords and prioritize these decoding paths. Then, SF is applied to improve the reliability of the multi-pronunciation hotword list. Finally, we utilize the \textbf{Uniform} reward scheme to guide Whisper’s decoding process toward hotword recognition by providing rewards for partial matches, thereby introducing a bias in favor of contextual biasing.
In Table~\ref{tab:proposed_librispeech}, with the reward scheme set to \textbf{Final}, comparing \textbf{S1} and the Baseline system, we observe a significant B-WER reduction (19\% for \textit{test-clean} and 22\% for \textit{test-other}), along with slightly better U-WER results. This suggests that a prefix-trie hotword implementation already provides substantial improvements in hotword recognition for Whisper. 
When we implement the multi-pronunciation hotword strategy without using SF (as in \textbf{S2} and \textbf{S3}), the U-WER performance degrades dramatically. At the same time, B-WER does not improve at all, showing that our pipeline for generating the multi-pronunciation hotwords brings us a large amount of noisy variations, leading to an incorrect rewarding paradigm.
However, when SF is applied to remove those faulty variations, comparing \textbf{S5} with S1, we can conclude that the strategy using multi-pronunciation hotwords with their original forms introduces further B-WER improvement by 10\% and 7\% for \textit{test-clean} and \textit{test-other}, respectively, while the U-WER remains essentially unchanged. Since the rare words only occupy around 10\% of the entire test sets, \textbf{S5} only achieves 2-3\% improvement on the overall WER. 
% This is primarily due to the multi-pronunciation variants with very short syllables, and variants with syllable count different from that of the corresponding hotword. Some variants consist of only one or two tokens, making them highly ambiguous and easily confused with frequent subword units, which leads to undesired hotword triggering and substitution. Moreover, since speech synthesis can be unstable for rare words and Whisper may hallucinate, the synthesized pronunciations sometimes deviate, and the syllable count differs from that of the original hotword. These mismatches collectively distort the decoding process and bias the output toward incorrect hotwords. After applying filtering and removing the above mismatches (rows with “\checkmark” in the filter column), we find that Multi-Pron-Hotword also yields modest gains compared with Vanilla Whisper-Large-v3, reducing B-WER to 8.73 \% on test-clean and to 15.68 \% on test-other. By combining these two, Multi-Pron-Hotword+Single-Pron achieves the lowest B-WER: 7.19 \% on test-clean and 12.60 \% on test-other. 
% U-WER also decreases, except for TTS biasing condition on test-clean, indicating that targeted hotword biasing does not impair general word recognition.

Using \textbf{Uniform} reward scheme, comparing \textbf{S6} with S1, U-WER and B-WER have identical improvement for both test sets, with around 3\% performance improvement on U-WER, and 23\% B-WER reduction. Also, we observe the same trends when comparing \textbf{S7} against S5. 
This indicates that guiding Whisper’s decoding process by rewarding step by step to favor rare words achieves even better performance.
Our final system, \textbf{S7}, integrates a prefix-trie contextual biasing module built on multi-pronunciation and the original form of rare words to the Baseline system, achieving 18\% and 21\% overall WER reduction, as well as 43\% and 44\% B-WER performance improvement, for \textit{test-clean} and \textit{test-other}, respectively.

We also compare our method against many prior contextual ASR approaches evaluated on the LibriSpeech dataset with biasing list from \cite{triebiasing} in Table \ref{tab:comparison_librispeech}. Despite utilizing a frozen Whisper backbone without fine-tuning or LLM assistance, our proposed method significantly outperforms several fine-tuned models in terms of B-WER under the same configuration of biasing list and distractor size ($N=1000$). 
% Importantly, our pipeline remains robust and scalable even with large biasing lists, demonstrating its practical suitability for real-world deployment.

\section{Conclusion}

We introduced a novel approach to enhancing zero-shot contextual ASR performance by leveraging synthetic speech for Whisper. Our method involves synthesizing diverse hotword speech via TTS engines and integrating these multi-pronunciation variants extracted from the Whisper model into a prefix-trie for shallow-fusion in beam-search decoding. Without any model fine-tuning or reliance on large language models, our system consistently improves hotword recognition performance on the LibriSpeech test sets with a hotword biasing list, achieving competitive B-WER while maintaining and even improving general ASR quality. 
% Crucially, we demonstrate that combining single-pronunciation-hotword and multi-pron-hotword variants, enforcing syllable filtering, and applying uniform per-token rewards are key to effective biasing. 
Our results surpass many fine-tuned contextual ASR systems, making the approach practical, low-cost, and easy to deploy, as it does not require any fine-tuning of the underlying ASR model.
Future work will expand the TTS augmentation pipeline by incorporating additional voice styles and more diverse TTS engines, including multilingual and expressive speech synthesis. This would enable coverage of broader pronunciation variability and better simulate real-world user scenarios across regions and accents.

\section*{Acknowledgment}
This research is supported by the RIE2025 Industry Alignment Fund - Industry Collaboration Projects (IAF-ICP) (Award I2301E0026), administered by A*STAR, as well as supported by Alibaba Group and NTU Singapore through Alibaba-NTU Global e-Sustainability CorpLab (ANGEL).
\newpage
\printbibliography

@article{advancesasr,
  title={Recent advances in end-to-end automatic speech recognition},
  author={Li, Jinyu and others},
  journal={APSIPA Transactions on Signal and Information Processing},
  volume={11},
  number={1},
  year={2022},
  publisher={Now Publishers, Inc.}
}

@inproceedings{seacoparaformer,
  title={Seaco-paraformer: A non-autoregressive asr system with flexible and effective hotword customization ability},
  author={Shi, Xian and Yang, Yexin and Li, Zerui and Chen, Yanni and Gao, Zhifu and Zhang, Shiliang},
  booktitle={ICASSP 2024-2024 IEEE International Conference on Acoustics, Speech and Signal Processing (ICASSP)},
  pages={10346--10350},
  year={2024},
  organization={IEEE}
}

@inproceedings{shallowfusion,
  title={Shallow-Fusion End-to-End Contextual Biasing.},
  author={Zhao, Ding and Sainath, Tara N and Rybach, David and Rondon, Pat and Bhatia, Deepti and Li, Bo and Pang, Ruoming},
  booktitle={Interspeech},
  pages={1418--1422},
  year={2019}
}

@inproceedings{wfstrescoring,
  title={Composition-based on-the-fly rescoring for salient n-gram biasing.},
  author={Hall, Keith B and Cho, Eunjoon and Allauzen, Cyril and Beaufays, Francoise and Coccaro, Noah and Nakajima, Kaisuke and Riley, Michael and Roark, Brian and Rybach, David and Zhang, Linda},
  booktitle={Interspeech},
  pages={1418--1422},
  year={2015}
}

@inproceedings{attentioncontextual,
  title={Deep context: end-to-end contextual speech recognition},
  author={Pundak, Golan and Sainath, Tara N and Prabhavalkar, Rohit and Kannan, Anjuli and Zhao, Ding},
  booktitle={2018 IEEE spoken language technology workshop (SLT)},
  pages={418--425},
  year={2018},
  organization={IEEE}
}

@article{triebiasing,
  title={Contextualized streaming end-to-end speech recognition with trie-based deep biasing and shallow fusion},
  author={Le, Duc and Jain, Mahaveer and Keren, Gil and Kim, Suyoun and Shi, Yangyang and Mahadeokar, Jay and Chan, Julian and Shangguan, Yuan and Fuegen, Christian and Kalinli, Ozlem and others},
  journal={arXiv preprint arXiv:2104.02194},
  year={2021}
}

@inproceedings{ctcllm,
  title={CTC-Assisted LLM-Based Contextual ASR},
  author={Yang, Guanrou and Ma, Ziyang and Gao, Zhifu and Zhang, Shiliang and Chen, Xie},
  booktitle={2024 IEEE Spoken Language Technology Workshop (SLT)},
  pages={126--131},
  year={2024},
  organization={IEEE}
}

@inproceedings{librispeech,
  title={Librispeech: an asr corpus based on public domain audio books},
  author={Panayotov, Vassil and Chen, Guoguo and Povey, Daniel and Khudanpur, Sanjeev},
  booktitle={2015 IEEE international conference on acoustics, speech and signal processing (ICASSP)},
  pages={5206--5210},
  year={2015},
  organization={IEEE}
}

@article{cppnet,
  title={Contextualized end-to-end speech recognition with contextual phrase prediction network},
  author={Huang, Kaixun and Zhang, Ao and Yang, Zhanheng and Guo, Pengcheng and Mu, Bingshen and Xu, Tianyi and Xie, Lei},
  journal={arXiv preprint arXiv:2305.12493},
  year={2023}
}

@inproceedings{biasphraseboosted,
  title={Contextualized automatic speech recognition with attention-based bias phrase boosted beam search},
  author={Sudo, Yui and Shakeel, Muhammad and Fukumoto, Yosuke and Peng, Yifan and Watanabe, Shinji},
  booktitle={ICASSP 2024-2024 IEEE International Conference on Acoustics, Speech and Signal Processing (ICASSP)},
  pages={10896--10900},
  year={2024},
  organization={IEEE}
}

@article{tcpgengnn,
  title={Tree-constrained pointer generator with graph neural network encodings for contextual speech recognition},
  author={Sun, Guangzhi and Zhang, Chao and Woodland, Philip C},
  journal={arXiv preprint arXiv:2207.00857},
  year={2022}
}

@inproceedings{gactc,
  title={Improving ASR contextual biasing with guided attention},
  author={Tang, Jiyang and Kim, Kwangyoun and Shon, Suwon and Wu, Felix and Sridhar, Prashant},
  booktitle={ICASSP 2024-2024 IEEE International Conference on Acoustics, Speech and Signal Processing (ICASSP)},
  pages={12096--12100},
  year={2024},
  organization={IEEE}
}

@inproceedings{phoneme,
  title={Phoneme-aware encoding for prefix-tree-based contextual ASR},
  author={Futami, Hayato and Tsunoo, Emiru and Kashiwagi, Yosuke and Ogawa, Hiroaki and Arora, Siddhant and Watanabe, Shinji},
  booktitle={ICASSP 2024-2024 IEEE International Conference on Acoustics, Speech and Signal Processing (ICASSP)},
  pages={10641--10645},
  year={2024},
  organization={IEEE}
}

@article{conformer,
  title={Conformer: Convolution-augmented transformer for speech recognition},
  author={Gulati, Anmol and Qin, James and Chiu, Chung-Cheng and Parmar, Niki and Zhang, Yu and Yu, Jiahui and Han, Wei and Wang, Shibo and Zhang, Zhengdong and Wu, Yonghui and others},
  journal={arXiv preprint arXiv:2005.08100},
  year={2020}
}

@article{attention,
  title={Attention is all you need},
  author={Vaswani, Ashish and Shazeer, Noam and Parmar, Niki and Uszkoreit, Jakob and Jones, Llion and Gomez, Aidan N and Kaiser, {\L}ukasz and Polosukhin, Illia},
  journal={Advances in neural information processing systems},
  volume={30},
  year={2017}
}

@article{rnn,
  title={Recurrent neural networks},
  author={Medsker, Larry R and Jain, Lakhmi and others},
  journal={Design and Applications},
  volume={5},
  number={64-67},
  pages={2},
  year={2001}
}

@article{f5tts,
  title={F5-tts: A fairytaler that fakes fluent and faithful speech with flow matching},
  author={Chen, Yushen and Niu, Zhikang and Ma, Ziyang and Deng, Keqi and Wang, Chunhui and Zhao, Jian and Yu, Kai and Chen, Xie},
  journal={arXiv preprint arXiv:2410.06885},
  year={2024}
}

@article{cosyvoice,
  title={Cosyvoice: A scalable multilingual zero-shot text-to-speech synthesizer based on supervised semantic tokens},
  author={Du, Zhihao and Chen, Qian and Zhang, Shiliang and Hu, Kai and Lu, Heng and Yang, Yexin and Hu, Hangrui and Zheng, Siqi and Gu, Yue and Ma, Ziyang and others},
  journal={arXiv preprint arXiv:2407.05407},
  year={2024}
}

@misc{gpt-sovits,
  title        = {{GPT-SoVITS}},
  year         = {2024},
  howpublished = {\url{https://github.com/RVC-Boss/GPT-SoVITS}}
}

@misc{voiceover_samples,
  title       = {{Voiceover Samples}},
  howpublished = {\url{https://www.voiceover-samples.com/languages/}},
  year         = {2025}
}

@misc{syllapy,
  title        = {syllapy},
  author       = {Michael Holtzscher},
  howpublished = {\url{https://pypi.org/project/syllapy/}},
  year         = {2025}
}

@inproceedings{TCPGen,
  title={Tree-constrained pointer generator for end-to-end contextual speech recognition},
  author={Sun, Guangzhi and Zhang, Chao and Woodland, Philip C},
  booktitle={2021 IEEE Automatic Speech Recognition and Understanding Workshop (ASRU)},
  pages={780--787},
  year={2021},
  organization={IEEE}
}

@article{whisper_gpt2,
  title={Can contextual biasing remain effective with Whisper and GPT-2?},
  author={Sun, Guangzhi and Zheng, Xianrui and Zhang, Chao and Woodland, Philip C},
  journal={arXiv preprint arXiv:2306.01942},
  year={2023}
}

\end{document}